\documentclass[letterpaper]{article} 
\usepackage[draft]{aaai2026}  
\usepackage{times}  
\usepackage{helvet}  
\usepackage{courier}  
\usepackage[hyphens]{url}  
\usepackage{graphicx} 
\usepackage{natbib}  
\usepackage{caption} 
\usepackage{multirow}
\usepackage{tabularx}
\usepackage{amsmath}
\usepackage{booktabs}
\usepackage{subcaption}
\newcolumntype{L}{>{\raggedright\arraybackslash}X}
\newcolumntype{C}{>{\centering\arraybackslash}X}
\newcolumntype{R}{>{\raggedleft\arraybackslash}X}

\title{Rethinking Graph-Based Document Classification:\\ Learning Data-Driven Structures Beyond Heuristic Approaches}

\author{
    Margarita Bugue\~no,
    Gerard de Melo
}
\affiliations{
    Hasso-Plattner Institute / University of Potsdam\\
    Potsdam, Germany\\
    \{margarita.bugueno, gerard.demelo\}@hpi.de
}

\begin{document}

\maketitle

\begin{abstract}
In document classification, graph-based models effectively capture document structure, overcoming sequence length limitations and enhancing contextual understanding. However, most existing graph document representations rely on heuristics, domain-specific rules, or expert knowledge. Unlike previous approaches, we propose a method to learn data-driven graph structures, eliminating the need for manual design and reducing domain dependence. Our approach constructs homogeneous weighted graphs with sentences as nodes, while edges are learned via a self-attention model that identifies dependencies between sentence pairs. A statistical filtering strategy aims to retain only strongly correlated sentences, improving graph quality while reducing the graph size. Experiments on three document classification datasets demonstrate that learned graphs consistently outperform heuristic-based graphs, achieving higher accuracy and $F_1$ score. Furthermore, our study demonstrates the effectiveness of the statistical filtering in improving classification robustness. These results highlight the potential of automatic graph generation over traditional heuristic approaches and open new directions for broader applications in NLP.
\end{abstract}

\begin{links}
     \link{Code}{https://github.com/Buguemar/AttnGraphs}
\end{links}

\section{Introduction}

Traditional vector-based text representation methods often struggle to effectively capture the structural information inherent in text, particularly when dealing with long documents. In contrast, graph-based representations have emerged as a powerful alternative, enabling the modeling of dependencies between textual units and leveraging their structure to better capture and differentiate local contexts within a document. Such representations have demonstrated promising results in document classification tasks \cite{zhang-etal-2020-every,wang2023text,gu2023enhancing,li2025cographnet}, with various graph construction strategies proposed to date.

However, existing graph-based approaches heavily rely on heuristics tailored to specific domains or tasks, requiring significant expert knowledge. As noted in a recent survey \cite{wang2023text}, graph structures in tasks like text classification are typically implicit, necessitating manual construction tailored to each application.
This dependency complicates the identification of their general effectiveness, as each construction method typically proves effective only within narrow, predefined scenarios \cite{bugueno-de-melo-2023-connecting,galke-scherp-2022-bag}. To address this limitation, a more robust and adaptable approach is needed to reduce the reliance on manually defined heuristics.

\begin{figure}[t]
\centering
\includegraphics[width=0.98\linewidth]{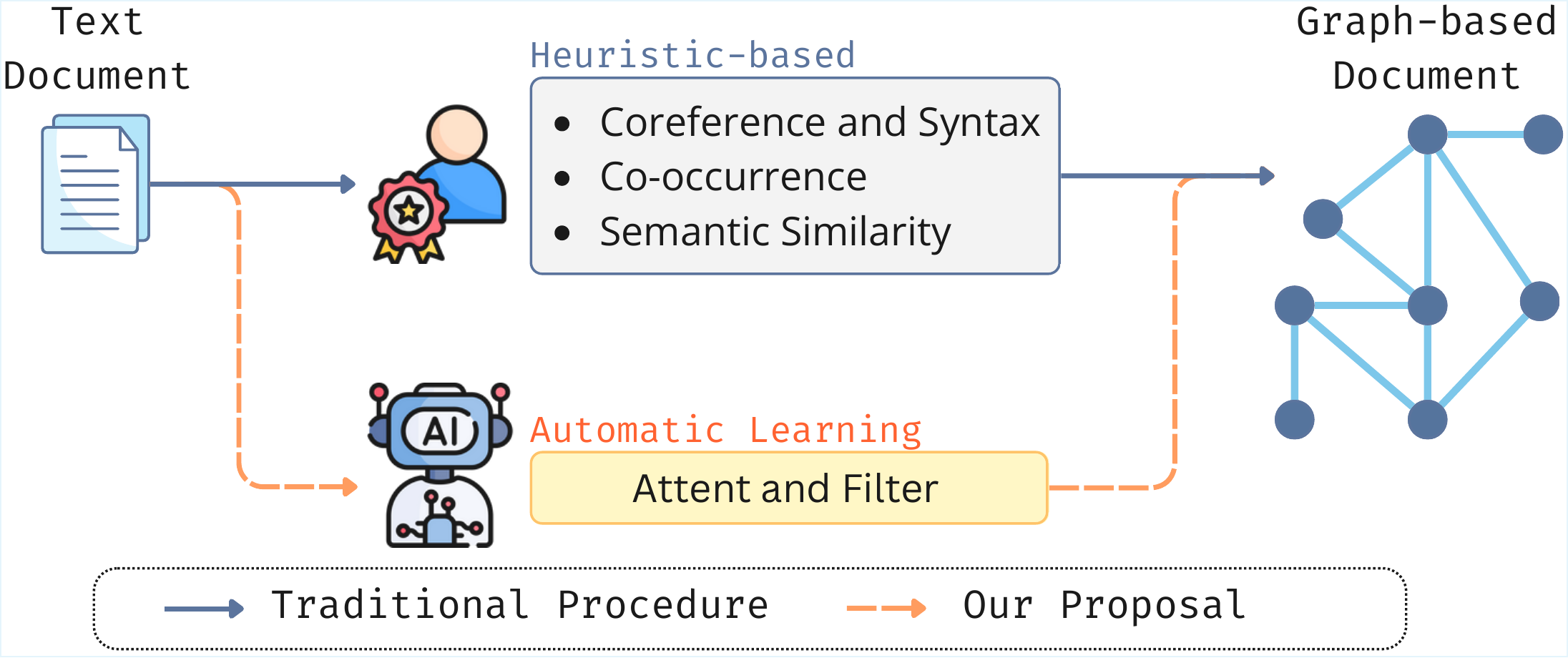}  
\caption{Unlike prior methods that rely on manually crafted heuristics and domain-specific rules for graph construction, our framework automatically learns graph structures from data. This removes the need for task-specific design and improves generalization across diverse applications.
}
\label{fig:scope}
\end{figure}

In this work, we propose a novel self-attention-based graph generation framework for document classification that, to our knowledge, is the first to automatically learn graph structures for document representations without relying on handcrafted heuristics, as traditional approaches do (see Figure \ref{fig:scope}). 
Our method constructs homogeneous graphs where nodes represent sentences within a document, and edges are determined by an attention model that learns relationships between sentence pairs. To retain only the most salient relationships, we apply a statistical filtering strategy to the learned attention weights, using either mean-bound or max-bound thresholds derived from the weight distribution.
 
To evaluate the effectiveness of our approach, we conducted experiments on three text classification datasets of varying lengths, comparing our learned graphs with four commonly used heuristic-based construction strategies: sentence order \cite{castillo2015author,bugueno-de-melo-2023-connecting}, window-based co-occurrence \cite{hassan-banea-2006-random,rousseau-etal-2015-text,zhang-etal-2020-every,li2025cographnet}, and semantic similarity under predefined thresholds \cite{li2025cographnet,mihalcea-tarau-2004-textrank,bugueno2024graphlss}.
Our findings reveal that attention-learned graphs consistently outperform heuristic-based graphs, with performance improvements becoming more pronounced as document length increases. 
Furthermore, our analysis shows that max-bound filtering is most effective for long documents, while mean-bound filtering performs best for medium-length documents. 

These results highlight the potential of automatically learned graphs over conventional heuristic approaches and open new directions for broader applications of graph-based document representations in NLP.

The key contributions of this paper are: 
\begin{itemize}
    \item A novel data-driven graph generation model: We introduce a self-attention-based approach that eliminates dependency on heuristics and domain expertise, significantly reducing the need for manual decisions. 
    \item Enhanced performance over heuristic-based graphs: Our proposed learned graphs demonstrate improvements of up to 4 points in accuracy and 4.3 points in $F_1$ score compared to traditional approaches.    
    \item Comprehensive evaluation and analysis: We conduct an extensive evaluation of two statistical filtering strategies applied to learned attention graphs, benchmarking their performance against four heuristic-based graph construction methods. This comparison is performed across multiple dimensions, including classification metrics, structural properties, and computational resource usage on three publicly available datasets.
\end{itemize}  

\section{Related Work}
\subsection{Predefined Graph Schemes}
\paragraph{Classic Approaches.} Numerous graph-based text representation approaches have been proposed for text classification, demonstrating the effectiveness of graph structures in capturing textual relationships. Early strategies focused on constructing graphs based on co-occurrence statistics and other linguistic patterns.

A common approach involved defining a fixed-size sliding window with words represented as nodes, and edges established between nodes if their corresponding words co-occur within a window of at most $N$ words \cite{mihalcea-tarau-2004-textrank,hassan-banea-2006-random,rousseau-etal-2015-text,zhang-etal-2020-every}. This simple yet effective construction captures local semantic associations. 

Another straightforward scheme involves sequence graphs, where edges reflect the original order of words in a document. While early implementations used edge weights corresponding to the frequency of consecutive word occurrences \cite{castillo2015author}, more recent work suggests that binarized edges are generally more effective in practice than weighted edges \cite{bugueno-de-melo-2023-connecting}.

\paragraph{Recent Approaches.} More sophisticated methods have been introduced, employing intricate structures to enhance textual modeling. One prominent approach is TextGCN \cite{Yao_Mao_Luo_2019}, which constructs a global heterogeneous graph consisting of word and document nodes, using Point-wise Mutual Information (PMI) for weighting word--word edges and TF-IDF for word--document links.
Conversely, TextLevelGCN \cite{huang-etal-2019-text} generates one graph for each text, where words serve as nodes (duplicated if they appear multiple times), and edges are defined between words within a sliding window, weighted by PMI.

Other studies integrate various heterogeneous contextual information to enrich graph representations, either by introducing topic nodes \cite{gu2023enhancing,cui-etal-2020-enhancing}, word and character n-gram nodes \cite{li-aletras-2022-improving}, or label nodes \cite{li2024graph}.  
Another strategy constructs an information graph composed of document keywords, entities, and titles \cite{ai2023multi}.

Furthermore, some approaches introduce multiple edge types while maintaining a single node type within the graph. Examples include constructing graphs with title, keyword, and event edges for document nodes \cite{ai2025contrastive}, as well as utilizing co-occurrence, syntactic dependency, and self-loop edges for graphs composed exclusively of word nodes \cite{wang2023text}.
An alternative strategy \cite{li2025cographnet} constructs separate heterogeneous graphs for words and sentences, which are subsequently fused during training. Word-graph edges are weighted based on the relative positioning of words within a specified window, while sentence-graph edges are derived from a combination of cosine similarity and positional bias.

\paragraph{Limitation.} Despite the advances made by these approaches, a fundamental limitation persists: they all rely on predefined domain knowledge to establish node and edge types. This dependency makes them heavily task- and domain-specific. To overcome this shortcoming, a learning-based approach for automatic graph structure discovery can eliminate the need for manual design and enhance generalizability and adaptability across diverse tasks and domains.

\subsection{Learning the Document Structure}
To the best of our knowledge, no previous method learns to generate a graph structure for document representation directly from the input text. Instead, all current strategies rely on domain-specific heuristics to construct graph structures representing textual documents. However, some related work has sought to enhance contextual document representations by integrating graph-based methods. 

\begin{figure*}[t!]
\centering
\includegraphics[width=0.98\textwidth]{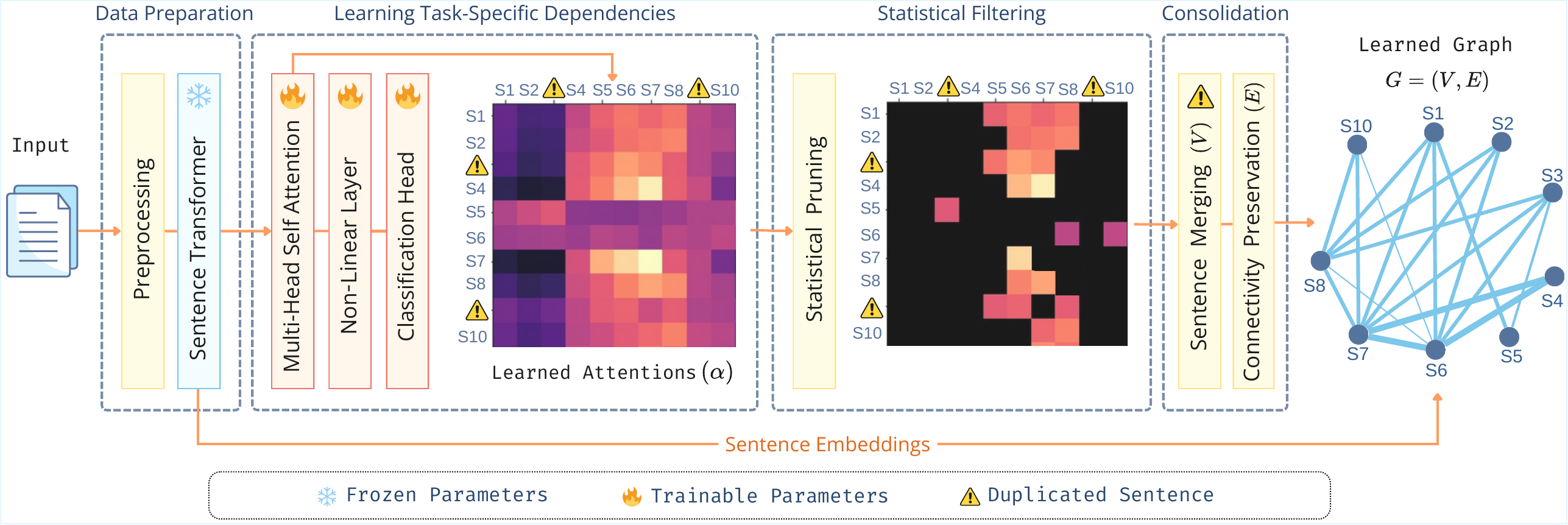} 
\caption{Overview of the proposed framework. ``Data preparation'', ``statistical filtering'', and ``consolidation'' are non-trainable steps, and the Sentence Transformer is used with frozen parameters. 
In the resulting graph $G$, edge width reflects the learned edge weights for the corresponding pair of nodes--thicker edges denote stronger dependencies.
}
\label{proposal}
\end{figure*}

The most relevant work \cite{xu-etal-2021-contrastive-document} proposes a framework that combines a Graph Attention Network (GAT) \cite{velivckovic2017graph} with a pre-trained Transformer encoder to learn document embeddings by exploiting the high-level semantic structure of documents. In this approach, documents are segmented into passages encoded using RoBERTa \cite{liu2019roberta}. The passages are organized into fully connected sub-graphs, each connected to a central document node represented by the average of all passage node embeddings. A GAT is then applied to capture multi-granularity document representations. 
Furthermore, the authors introduce a document-level contrastive learning strategy to pre-train their model and  enhance representation learning. While effective, this method does not learn the underlying graph topology. 
Rather, it identifies relevant passages for document representation by leveraging the document structure through a predefined GAT architecture.

Recent studies have increasingly focused on integrating graph structures with pre-trained language models to enhance document representation learning, recognizing the potential of combining representations that capture local node interactions with powerful contextual encoders. One such study \cite{huang-etal-2022-contexting} proposes a unified model combining Graph Neural Network (GNN) models and BERT \cite{devlin-etal-2019-bert} to learn contextual inductive document representations. The method employs a sub-word graph to emphasize fine-grained syntactic relationships, thereby mitigating the overemphasis on content-specific word usages. Similarly, another study \cite{onan2023hierarchical} introduces a hierarchical graph-based framework for text classification, where BERT encodes contextual information for each graph node, resulting in improved representation learning and classification performance.

A recent method for multi-label document classification \cite{liu2025all} exclusively employs attention mechanisms. It constructs text and label embeddings using the pre-trained model XLNet \cite{yang2019xlnet} and generates a graph structure based on label co-occurrence to preserve label correlation information. A graph attention mechanism learns label dependencies from the graph structure and semantic relationships among labels. Moreover, a class-specific attention module creates distinct feature spaces for each category label, while a self-attention mechanism enhances the model’s ability to capture contextual dependencies within the text.

Although there are numerous graph-based strategies in the literature, the challenge of automatically learning graph topologies for document representation directly from raw text remains largely unexplored. 
Moreover, recent work emphasizes the effectiveness of integrating attention mechanisms and pre-trained language models for building robust and adaptive graph-based document representations. 
It also highlights the limitations of traditional heuristic-driven graph construction methods, 
particularly in handling diverse domains and coping with modern document processing requirements such as large-scale data, capturing long-range dependencies, and dealing with noisy and imbalanced data.

\section{Learning Data-Driven Document Graphs}

We introduce a novel approach for learning data-driven graph structures, eliminating the reliance on manual design and minimizing domain dependency.
Our methodology builds upon insights from previous work \cite{xu-etal-2021-contrastive-document,liu2025all}, highlighting the capabilities of pre-trained language models and attention mechanisms for capturing contextual relationships.
To this end, our framework constructs homogeneous weighted graphs, where sentences are represented as nodes, and edges are learned via a self-attention model that captures dependencies between sentence pairs. Specifically, given a document $D$, our approach generates graphs $G=(V, E)$, where $V= \{s_i, s_2, \ldots, s_n \}$ with $n$ denoting the number of sentences in $D$. 
The edge set $E$ is defined as $\{\alpha_{ij} \mid \alpha_{ij} \geq \tau_i \}$ for every sentence pair $(i,j)$ in $D$, where $\tau_i$ is a pre-calculated attention threshold for every sentence $s_i \in D$.

The decision to use sentences as nodes is motivated not only by previous research demonstrating their effectiveness in delineating the logical structure of documents but also by their scalability for long documents. 
Furthermore, we generate homogeneous rather than heterogeneous graphs, as the latter are far more resource-intensive and highly rely on external tools \cite{sahu-etal-2019-inter,wang2023text,ai2025contrastive}. 
Previous work also suggests that simpler graph constructions often yield better results compared to more specialized graphs \cite{bugueno-de-melo-2023-connecting}.

Following the learning of attention weights for all sentence pairs in the document, a statistical filtering mechanism is applied. This filter establishes a minimum threshold for each row $i$ ($\tau_i$) in the attention matrix, ensuring that only strongly correlated sentence pairs ($\alpha_{ij}$) are retained, as well as ensuring connected graphs, i.e., all vertices in the graph are reachable. Thereby, we enhance the graph quality while reducing the graph complexity. 
The overall framework of our proposed model is illustrated in Figure \ref{proposal}.
A detailed step-by-step description follows.

\subsection{Data Preparation}

Prior to training the self-attention model, it is essential to define the units that will serve as nodes within the learned graphs, namely, the sentences. This process includes a thorough data-cleaning procedure followed by sentence tokenization.\footnote{Implemented using the NLTK library in Python.}
To prevent the graph size from growing excessively and ensure computational efficiency, sentences containing fewer than five words are merged with the preceding one. 
This preprocessing step helps maintain meaningful sentence representations while reducing unnecessary complexity in graph construction.

\subsection{Learning Task-Specific Dependencies}
The proposed approach for document classification begins by segmenting the input document $D$ into a sequence of its constituent sentences $s_1, s_2, \dots, s_n$, where $n$ denotes the total number of sentences in the document after preprocessing. 
This segmentation allows the model to capture sentence-level dependencies that are essential to accurately modeling the overall structure of the document graph.

To obtain vector representations, we map each sentence $s_i$ into a fixed-dimensional embedding $x_i \in \mathrm{R}^d$, using a pre-trained Sentence Transformer, with $d=384$ in our experiments. 
The resulting set of embeddings $x_i, x_2, \dots, x_n$ serves as the input representation of the document, effectively transforming the textual data into vector representations for further processing.

Building upon these representations, a multi-head self-attention model is trained to learn inter-sentence dependencies. The architecture comprises a multi-head attention mechanism, followed by a non-linear layer using ReLU, and concludes with a classification head designed to perform document classification across the available classes. 
Inspired by promising results in prior work \cite{wortsman2023replacing}, we substitute the conventional softmax activation function used during the scaled dot-product attention computation with a ReLU activation normalized by the document sequence length. This modification seeks to provide a more efficient and effective attention mechanism, an approach that has also demonstrated empirical success in recent studies \cite{bai2023transformers,zhao2024mobilediffusion}.
The learned attention matrix per document is given by $\alpha_{ij}$ for pair sentences $s_i$ and $s_j$.
Further details are provided in the Section \ref{setup}.

\subsection{Statistical Filtering}
To enhance the relevance of the attention weights produced by the multi-head self-attention model, we apply a statistical filtering step that selectively discards weak dependencies while retaining only those sentence pairs ($\alpha_{ij}$) deemed salient for the document classification task. 
This process effectively transforms attention weights into graph edges representing meaningful relationships between sentences. Filtering is conducted row-wise to ensure the generation of connected graphs, establishing at least one edge for each document sentence. Additionally, the filtering also accounts for self-loops, discarding such edges from the matrix.
Two alternative filtering strategies are introduced.

\paragraph{\textbf{Mean-bound}.} This approach computes the average attention score for each sentence $s_i$ across all other sentences within the document and derives a minimum attention threshold incorporating a predefined tolerance degree $\delta$. The threshold is given by:
\begin{equation}
  \tau_i =  \frac{1}{n}\sum_{j=1}^n \alpha_{ij} + \delta \cdot \text{std}(\alpha_i) \ ,
  \label{mean}
\end{equation}
where $\text{std}(\alpha_i)$ is the standard deviation of the $i$-row of the learned attention matrix. This threshold is slightly greater than the mean, which reduces the tolerance level and decreases the number of retained entries in the attention matrix, thereby ensuring that only the most relevant dependencies are preserved.

\paragraph{\textbf{Max-bound}.} This strategy focuses on top-ranked dependencies, retaining attention scores proximate to the maximum observed value within each row, i.e., for each sentence $s_i$ in the document. The threshold is calculated as:
\begin{equation}
  \tau_i =  \text{max}_j(\alpha_{ij}) - \delta \cdot \text{std}(\alpha_i) \ ,
  \label{max} 
\end{equation}
where $\text{std}(\alpha_i)$, as in Equation \ref{mean}, is the standard deviation of the $i$-row of the learned attention matrix.
Notably, we increase the tolerance for preserving entries around the peak attention score for each row, yielding a more aggressive pruning criterion.

\subsection{Consolidation}
Following the statistical filtering process, the resulting matrix is interpreted as the adjacency matrix of the learned graph. To ensure structural coherence, two operations are performed to account for special edge cases.

\paragraph{Sentence Merging.} When identical sentences are present at different positions within $D$, their corresponding edges in the adjacency matrix are unified to maintain the integrity of the graph representation and better reflect the semantic structure of the document. For instance, if $D = \{ s_1, s_2, s_3, s_4, s_5, s_6 \}$, with $s_2 = s_5$, the edges associated with $s_2$ and $s_5$ are merged, resulting in a reduced graph with five unique sentence nodes.
This step ensures consistency and avoids redundancy, adjusting the set of effective sentence nodes in the final learned graph. 

\paragraph{Connectivity Preservation.} 
Disconnected graphs need to be avoided. A typical scenario arises when 
there is no plausible edge for the row $\alpha_i$ ($s_i$) after statistical filtering, which fails to establish meaningful connections with other sentences.
To resolve this issue, additional edges are introduced by connecting the sentence node $s_i$ to its immediately preceding and subsequent sentence nodes. 
The original attention weight associated with the self-loop $\alpha_{ii}$ is evenly distributed between these newly established edges, which guarantees graph connectivity while preserving the original attention-based weighting scheme.

Finally, the learned graph $G=(V, E)$ consists of unique sentence nodes $V \in D$, encoded via Sentence Transformer embeddings, and undirected, attention-weighted edges $E$ that effectively capture the document structure.

\section{Experiments}
To study the merits of our learned graphs for document representation, we conducted comprehensive experiments on three publicly available text classification datasets (see Table \ref{datasets}), covering documents of varying lengths and domains. 
For each task, we compare our learned graphs against five heuristic-based graph construction schemes by training a GAT model under consistent experimental conditions. 

\subsection{Datasets}
We assess the generalizability of our model across balanced and unbalanced scenarios, focusing on topic classification and hyperpartisan news detection in three different settings: medium-length news articles, long news articles, and very long scientific papers.

\begin{itemize}
    \item \textbf{BBC News}\footnote{\url{http://derekgreene.com/bbc/}} \cite{greene2006practical}: A moderately imbalanced collection of 2,225 English documents from the BBC News website (2004--2005) in the areas of business, entertainment, politics, sport, and technology. After duplicate removal, we partition the data into training (1,547), validation (177), and test (443) sets.
    \item \textbf{Hyperpartisan News Detection (HND)}\footnote{\url{https://zenodo.org/records/5776081}} \cite{kiesel-etal-2019-semeval}: English news articles labeled according to whether they show blind or unreasoned allegiance to a single political party or entity, or not.  Although it comprises two parts, \texttt{byarticle} and \texttt{bypublisher}, we use the first one with 645 training and 625 test samples.  
    \item \textbf{arXiv}\footnote{\url{https://huggingface.co/datasets/ccdv/arxiv-classification}} \cite{he2019long}: A collection of 33,388 long scientific papers in physics, mathematics, computer science, and biology sourced from the arXiv. The 11-class dataset exhibits slight imbalance and is divided into 3 splits: train (28,000), validation (2,500), and test (2,500).
\end{itemize}
For all experiments, we remove duplicate samples and perform an 80\%/20\% training--test split for BBC News, as a predefined test set was not available. Additionally, we randomly select 10\% of the training data for validation.

\begin{table}[!t]
\centering
\begin{tabular}{p{1.7cm}p{3.5cm}>{\raggedleft\arraybackslash}p{0.7cm}>{\raggedleft\arraybackslash}p{0.7cm}} 
\toprule
\textbf{Dataset} & \textbf{Avg.\ Length}      & \textbf{K} & \textbf{IR}\\ \hline
BBC News         & 438 words (19 sent.)  & 5          & 4:5  \\ 
HND              & 912 words (21 sent.)  & 2          & 1:2  \\
arXiv            & 10,554 words (539 sent.) & 11       & 1:2 \\
\bottomrule
\end{tabular}
\caption{Statistics of datasets. This includes the average document length in terms of words and sentences, the number of classes (K), and the imbalance rate between the minority and majority classes (IR).}
\label{datasets}
\end{table}

\subsection{Heuristic-based Graphs}
We evaluate the performance of our learned graphs against five widely adopted heuristic-based homogeneous graph construction strategies. 
In all cases, graph nodes correspond to the unique sentences within a document $D$. Specifically, we consider the following methods: 
\begin{itemize}
    \item \textbf{Complete Graph}: It serves as a fundamental baseline, where each sentence node is fully connected to all others using unweighted edges, forming a complete graph.
    \item \textbf{Sentence Order}: It constructs edges based on the natural order of sentence occurrence within the document. Undirected binary edges (0/1) are established without incorporating attributes or edge weights. This simplistic approach solely captures the sequential structure of the text.
    \item \textbf{Window-based Co-Occurrence}: Undirected edges are established between sentence nodes if they co-occur within a fixed sliding window of size 3. Therefore, each sentence node is connected to its two preceding and two subsequent sentences. Notably, this construction can be considered a generalization of the sentence order-based graph by capturing broader contextual dependencies.
    \item \textbf{Semantic Similarity with Mean Threshold}: Weighted edges are defined based on a cosine similarity threshold applied to the corresponding sentence embeddings. The threshold is determined by following the procedure described in Equation \ref{mean}, providing a fair comparison against our learned graphs.  
    \item \textbf{Semantic Similarity with Max Threshold}: Similar to the mean threshold-based construction, but using the cosine similarity thresholding procedure outlined in Equation \ref{max}. As a result, sparser graphs are expected, retaining only the most prominent connections.
\end{itemize}

\begin{table*}[!t]
    \centering    
    \begin{tabularx}{\textwidth}{p{3.5cm}RR@{\hskip 0.3cm}RRR@{\hskip 0.3cm}R}
        \toprule
        \textbf{Graph Scheme} & \textbf{Accuracy} & \textbf{$\mathbf{F_1}$-ma} & \textbf{$\mathbf{|V|}$} & \textbf{$\mathbf{|E|}$} & \textbf{Degree} & \textbf{Disk} \\
        \hline
        2 L - 64 U & \multicolumn{6}{c}{\textit{BBC News}} \\
        \hline
        complete graph &  $\mathbf{99.9}$ & $\mathbf{99.9}$ & 19.30 & 481.69 & 18.30 & 105 MB \\
        sentence order & 99.7 & 99.7 & 19.30 & 36.61 & 1.87 & 74 MB\\
        window co-occurrence & 99.8 & 99.8 & 19.30 & 71.21 & 3.62 & 76 MB \\
        mean semantic similarity & 99.4 & 99.3 & 19.30 & 159.68 & 5.40 & 84 MB\\
        max semantic similarity & 99.7 & 99.7 & 19.30 & 36.66 & 1.88 & 74 MB\\
        \hline
        learned mean-bound & $\mathbf{99.9}$ & $\mathbf{99.9}$ & 19.30 & 245.76 & 9.52 & 90 MB \\ 
        learned max-bound & 99.6 & 99.6 & 19.30 & 66.33 & 3.39 & 77 MB\\        
        \hline
        3 L - 64 U & \multicolumn{6}{c}{\textit{Hyperpartisan News Detection}} \\
        \hline 
        complete graph &  94.6 & 94.5 & 19.48 & 710.90 & 18.49 & 70 MB \\
        sentence order & 92.6 & 92.6 & 19.48 & 37.00 & 1.78 & 43 MB\\
        window co-occurrence & 92.1 & 92.1 & 19.48 & 71.98 & 3.36 & 44 MB\\
        mean semantic similarity & 91.2 & 91.1 & 19.48& 254.84 & 6.00 & 53 MB \\
        max semantic similarity & 92.8 & 92.8 & 19.48 & 36.93 & 1.79 & 43 MB\\
        \hline
        learned mean-bound & $\mathbf{95.0}$ & $\mathbf{94.9}$ & 19.48 & 329.60 & 8.86 & 56 MB\\ 
        learned max-bound & 92.6 & 92.6 & 19.48 & 57.38 & 2.79 & 44 MB\\
        \hline
        3 L - 64 U &  \multicolumn{6}{c}{\textit{arXiv}} \\
        \hline 
        sentence order & 87.3 & 86.7 & 510.33 & 1,035.05 & 2.02 & 25 GB\\
        window co-occurrence & 87.9 & 87.4 & 510.33 & 1,068.25 & 4.04 & 26 GB\\
        max semantic similarity & 87.8 & 87.4 & 510.33 & 1,241.94 & 2.28 & 26 GB\\ 
        \hline
        learned max-bound & $\mathbf{91.9}$ & $\mathbf{91.7}$ & 510.33 & 1,092.20 & 2.16 & 25 GB\\  
        \bottomrule        
    \end{tabularx}
        \caption{Structural features and classification results of heuristic-based and learned graphs across datasets. Metrics include accuracy, macro-averaged $F_1$ score, average number of nodes, edges, and degree, and total disk usage. Results for \textit{complete graph}, \textit{mean semantic similarity}, and \textit{mean-bound learned} are omitted on arXiv due to prohibitive computational overhead.}
    \label{tab:mainresults}
\end{table*}

\subsection{Experimental Setup}
\label{setup}

To address particularly long documents, such as those in the arXiv dataset, we employed a cut-off mechanism by defining dataset-specific maximum sequence lengths. For BBC News and HND, we preserved full documents with limits of 185 and 136 sentences, respectively. For arXiv, the maximum sequence length was 1,800 sentences. This threshold was deliberately set high to minimize information loss, resulting in truncation for fewer than 1.5\% of documents.

To obtain sentence embeddings, we utilized the pre-trained Sentence Transformer model \texttt{paraphrase-MiniLM-L6-v2}\footnote{\url{https://huggingface.co/sentence-transformers/paraphrase-MiniLM-L6-v2}}. Notably, the attention model architecture comprises a single layer of multi-head self-attention; however, additional experiments with a two-layer architecture are reported in Table \ref{tab:robustness}, Section \ref{sec:results}. The tolerance degree $\delta$ in Equation \ref{mean} and Equation \ref{max} is set to 0.5 throughout all experiments. 

\paragraph{\textbf{Self-Attention Model}} 
The multi-head self-attention models employed four attention heads and a batch size of 32 samples. The models were trained for a maximum of 20 epochs using Adam optimization \cite{kingma2014adam} with an initial learning rate of 0.001. 
Training was interrupted if the validation macro-averaged $F_1$ score did not improve for five consecutive epochs.

In our implementation, the resulting learned document graphs are stored as PyTorch Geometric objects. While alternative approaches construct graphs on the fly, we precompute and save the graphs, incurring the graph-creation cost only once. This optimization significantly reduces computational overhead by eliminating the need for graph reconstruction across epochs and model variations.

\paragraph{\textbf{Graph Attention Network (GAT)}} 
We assessed the performance of GAT architectures with 1 to 3 hidden layers and node embedding sizes in $\{64, 128, 256\}$. Dropout was applied after each convolutional layer with a retention probability of 0.8, and average pooling was used for node-level aggregation. The resulting representations were passed through a softmax layer for final classification. All GAT experiments were implemented in PyTorch Geometric.

Training was conducted for a maximum of 50 epochs with a batch size of 64, utilizing the Adam optimizer \cite{kingma2014adam} with an initial learning rate of 0.001. 
Early stopping based on the validation macro-averaged $F_1$ score was applied as in the self-attention model.

\section{Results}
\label{sec:results}
The main results are presented in Table \ref{tab:mainresults}. These correspond to the average obtained from 5 independent runs. 
All experiments are based on PyTorch Geometric and conducted on an NVIDIA GeForce RTX3050.

\paragraph{\textbf{Quality of the Results}}
The proposed learned graphs consistently outperform heuristic-based graph construction strategies across all three evaluated datasets. 
While the performance gains on BBC News are marginal, the advantages of our approach become increasingly pronounced as the document length increases. Notably, although the complete graph baseline reported the same classification performance as our learned mean-bound graphs, it does so at the expense of nearly twice the number of edges, needing an additional 15 MB for storage. 
On the HND dataset, our learned mean-bound graphs surpass the best-performing heuristic-based approach--max semantic similarity--by up to 2.1 $F_1$ points. This improvement is even more pronounced on the arXiv dataset, achieving a gain of 4.3 $F_1$ points, emphasizing the effectiveness of our method in capturing document structure for classification tasks. 

As stated in Table \ref{tab:mainresults}, due to the varying lengths of the datasets under study, the GAT architectures are adapted accordingly. For BBC News, which comprises shorter documents, the best-performing model consists of a 2-layer GAT with 64 hidden units. In contrast, a deeper architecture (three layers, 64 units) is employed for HND and arXiv, which contain substantially longer documents. Such architecture provides a greater capacity to capture the complex semantic relationships present in lengthy documents.
Due to their high computational and memory demands, the heuristic-based complete graph and mean semantic similarity graph variants, as well as our learned mean-bound graphs, are excluded from experiments on arXiv. These resource-intensive requirements become particularly prohibitive when processing extremely long documents. 

\begin{figure}[t!]
\centering
  \begin{subfigure}{0.99\columnwidth}
    \includegraphics[width=\linewidth]{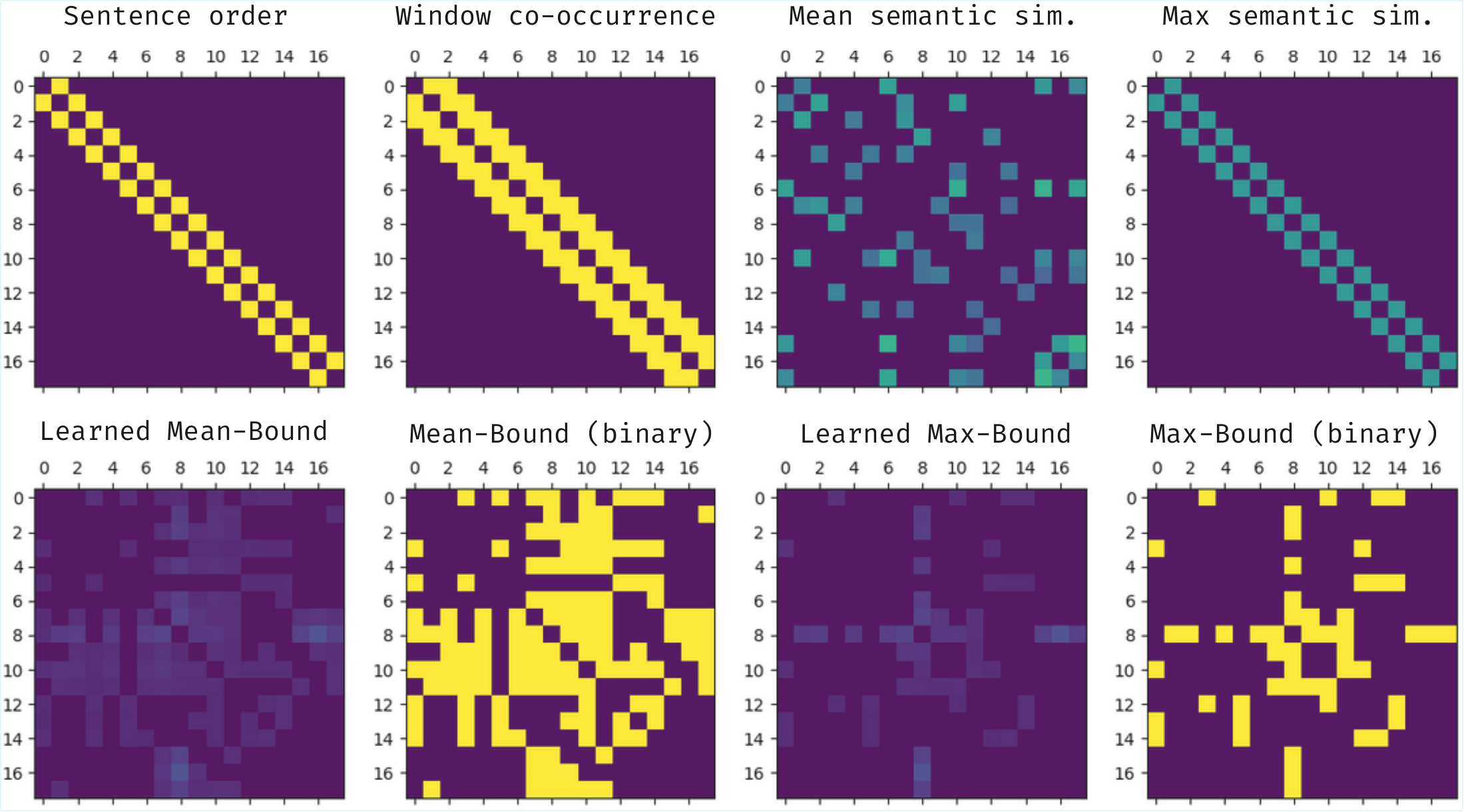}
    \caption{A random BBC News sample.} 
    \label{fig:bbc}
  \end{subfigure}%
  \\ \vspace{0.2cm}
  \begin{subfigure}{0.99\columnwidth}
    \includegraphics[width=\linewidth]{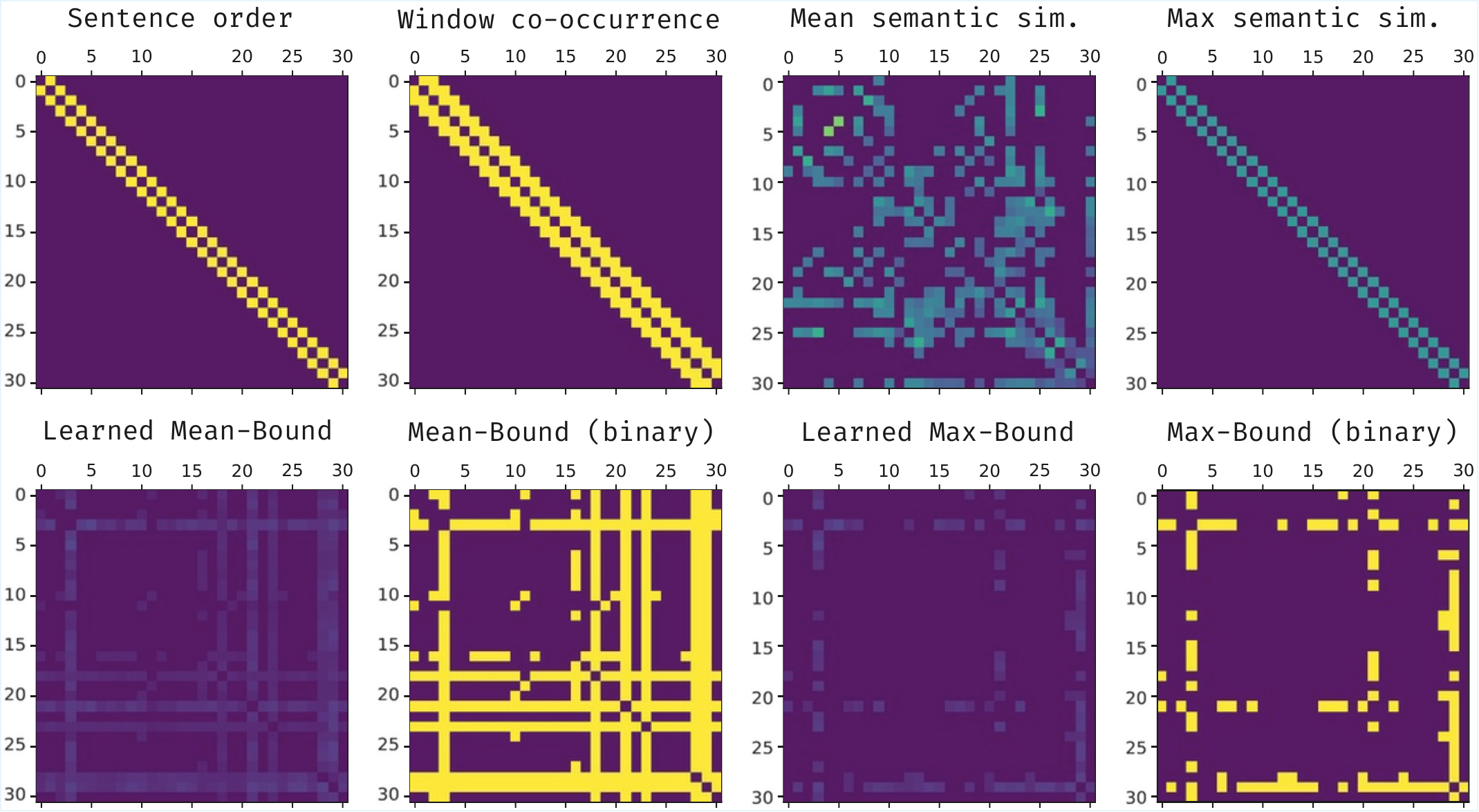}
    \caption{A random HND sample.} 
    \label{fig:hnd}
  \end{subfigure}%
  \\ \vspace{0.2cm}
  \begin{subfigure}{0.99\columnwidth}
    \includegraphics[width=\linewidth]{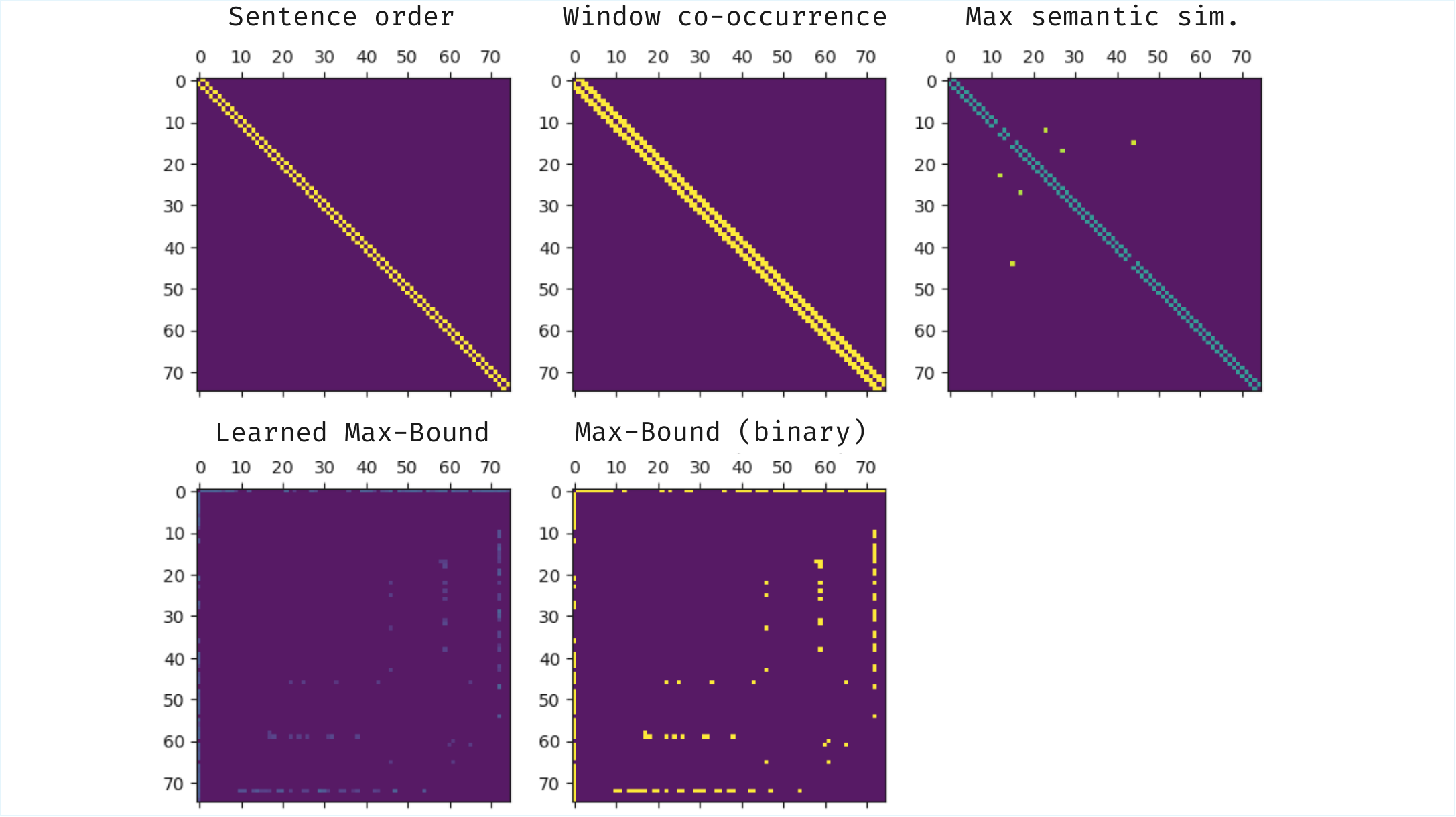}
    \caption{A random arXiv sample.} 
    \label{fig:arxiv}
  \end{subfigure}

\caption{Adjency matrix comparison for each graph scheme on a randomly sampled document from each dataset.} 
\label{fig:adjency_all}
\end{figure}

\paragraph{\textbf{Graph Structure Analysis}}
A key advantage of our proposal is its ability to capture global contextual dependencies within a document. Unlike heuristic-based graph constructions, which rely on a predefined window size and are therefore constrained to local sentence relationships, our approach allows edges between relevant but distant sentences, considering all sentences simultaneously and thereby enhancing the expressiveness of the learned structure. 

Despite comparable storage requirements between our learned mean-bound graphs and the heuristic-based mean semantic similarity graphs, the disparity in performance metrics is substantial. Both variants report the highest average degree across all evaluated datasets--in addition to complete graph--, yet the superior performance of our approach cannot be attributed to graph density. Instead, results demonstrate that the edges learned by our model effectively capture the underlying semantics and structural relationships present within the documents.
Furthermore, for the arXiv dataset, the most effective heuristic-based graphs (i.e., window co-occurrence and max semantic similarity) exhibit a higher average degree than our learned max-bound graphs, further underscoring the robustness of our approach.
Visualizations of adjacency matrices (Figure \ref{fig:adjency_all}) underscore the importance of capturing comprehensive document structures. The figures highlight the significance of both the initial and final sentences of the document in achieving accurate classification, particularly in long-form documents like those in arXiv.
For clarity, we include binarized versions of the learned adjacency matrices, as they typically exhibit lower edge weight values than heuristic-based graphs.

\begin{table}[!t]
    \centering
    \begin{tabularx}{\linewidth}{p{1.7cm}|CCCr}
    \toprule
         \textbf{Dataset} & \textbf{L} & \textbf{Acc} & \textbf{$F_1$} & \textbf{$F_1$ score per class}\\
         \hline
         
         \multirow{10}{\hsize{}}{BBC News} & \multirow{5}{*}{1} & \multirow{5}{*}{95.5} & \multirow{5}{*}{95.3} & sport: 98.7 \\
                                          &   &      &      & entertainment: 95.2\\
                                          &   &      &      & business: 94.1\\
                                          &   &      &      & tech: 96.8\\
                                          &   &      &      & politics: 91.4\\
                                          \cline{2-5}
                                   & \multirow{5}{*}{2} & \multirow{5}{*}{95.5} & \multirow{5}{*}{95.2} & sport: 99.1\\
                                   &   &    &   & entertainment: 92.9\\
                                   &   &    &   & business: 93.5\\
                                   &   &    &   & tech: 97.4\\
                                   &   &    &   & politics: 93.2\\ \hline
         \multirow{4}{*}{HND}      & \multirow{2}{*}{1} & \multirow{2}{*}{76.4} & \multirow{2}{*}{76.4} & non-hyperpartisan: 75.5\\
                                   &    &   &   & hyperpartisan: 77.3\\
                                   \cline{2-5}
                                   & \multirow{2}{*}{2} & \multirow{2}{*}{76.4} & \multirow{2}{*}{76.3} & non-hyperpartisan: 76.4\\
                                   &    &   &   & hyperpartisan: 76.2\\
    \bottomrule         
    \end{tabularx}
    \caption{Classification results obtained by a 1- and a 2-layer (L) self-attention model. Acc and $F_1$ stand for accuracy and macro-averaged $F_1$ score, respectively.}
    \label{tab:robustness}
\end{table}

\paragraph{\textbf{Robustness}}
\label{par:robustness}
As Table \ref{tab:robustness} shows, our method demonstrates strong robustness across model architectures. Even shallow self-attention models induce strong document representations. Notably, it is essential for the learned attention weights to exhibit sparsity, which is critical for effectively identifying potential edges throughout the document. This sparsity facilitates the subsequent training of GAT models by efficiently exploring and leveraging the local neighborhood structure within the learned graph, enhancing its capacity to capture meaningful relationships within the document.

\section{Conclusion}
We present a novel framework for learning data-driven graph structures for document representation, effectively eliminating the need for manual task-specific graph design and reducing dependency on expert knowledge and domains. Comprehensive experiments on three document classification datasets demonstrate that our learned graphs consistently surpass traditional heuristic-based graph constructions concerning accuracy and $F_1$ score, capturing the long-range and non-sequential dependencies that sentences can have among themselves. These findings underscore the efficacy of automatic graph generation, suggesting promising directions for broader applications. Future work will explore alternative filtering strategies, additional tasks, and examine hierarchical methods for learning heterogeneous graphs.

\bibliography{bibliography}

\begin{thebibliography}{52}
\providecommand{\natexlab}[1]{#1}

\bibitem[{Ai et~al.(2025)Ai, Li, Wang, Wei, Meng, and Li}]{ai2025contrastive}
Ai, W.; Li, J.; Wang, Z.; Wei, Y.; Meng, T.; and Li, K. 2025.
\newblock Contrastive multi-graph learning with neighbor hierarchical sifting for semi-supervised text classification.
\newblock \emph{Expert Systems with Applications}, 266.

\bibitem[{Ai et~al.(2023)Ai, Wang, Shao, Meng, and Li}]{ai2023multi}
Ai, W.; Wang, Z.; Shao, H.; Meng, T.; and Li, K. 2023.
\newblock A multi-semantic passing framework for semi-supervised long text classification.
\newblock \emph{Applied Intelligence}, 53(17): 20174--20190.

\bibitem[{Bai et~al.(2023)Bai, Chen, Wang, Xiong, and Mei}]{bai2023transformers}
Bai, Y.; Chen, F.; Wang, H.; Xiong, C.; and Mei, S. 2023.
\newblock Transformers as statisticians: Provable in-context learning with in-context algorithm selection.
\newblock \emph{Advances in Neural Information Processing Systems}, 36: 57125--57211.

\bibitem[{Beltagy, Peters, and Cohan(2020)}]{beltagy2020longformer}
Beltagy, I.; Peters, M.~E.; and Cohan, A. 2020.
\newblock Longformer: The long-document transformer.
\newblock \emph{arXiv preprint arXiv:2004.05150}.

\bibitem[{Bugue{\~n}o and de~Melo(2023)}]{bugueno-de-melo-2023-connecting}
Bugue{\~n}o, M.; and de~Melo, G. 2023.
\newblock Connecting the Dots: What Graph-Based Text Representations Work Best for Text Classification using Graph Neural Networks?
\newblock In \emph{Findings of the Association for Computational Linguistics: EMNLP 2023}, 8943--8960.

\bibitem[{Bugue{\~n}o, Hamdan, and de~Melo(2024)}]{bugueno2024graphlss}
Bugue{\~n}o, M.; Hamdan, H.~A.; and de~Melo, G. 2024.
\newblock GraphLSS: Integrating Lexical, Structural, and Semantic Features for Long Document Extractive Summarization.
\newblock \emph{arXiv preprint arXiv:2410.21315}.

\bibitem[{Castillo et~al.(2015)Castillo, Cervantes, Vilarino, and B{\'a}ez-L{\'o}pez}]{castillo2015author}
Castillo, E.; Cervantes, O.; Vilarino, D.; and B{\'a}ez-L{\'o}pez, D. 2015.
\newblock Author verification using a graph-based representation.
\newblock \emph{International Journal of Computer Applications}, 123(14): 1--8.

\bibitem[{Condevaux and Harispe(2023)}]{10.1007/978-3-031-33374-3_35}
Condevaux, C.; and Harispe, S. 2023.
\newblock LSG Attention: Extrapolation of Pretrained Transformers to Long Sequences.
\newblock In Kashima, H.; Ide, T.; and Peng, W.-C., eds., \emph{Advances in Knowledge Discovery and Data Mining}, 443--454. Cham: Springer Nature Switzerland.
\newblock ISBN 978-3-031-33374-3.

\bibitem[{Cui, Hu, and Liu(2020)}]{cui-etal-2020-enhancing}
Cui, P.; Hu, L.; and Liu, Y. 2020.
\newblock Enhancing Extractive Text Summarization with Topic-Aware Graph Neural Networks.
\newblock In \emph{Proceedings of the 28th International Conference on Computational Linguistics}, 5360--5371.

\bibitem[{Devlin et~al.(2018)Devlin, Chang, Lee, and Toutanova}]{devlin2018bert}
Devlin, J.; Chang, M.-W.; Lee, K.; and Toutanova, K. 2018.
\newblock Bert: Bidirectional encoder representations from transformers.
\newblock \emph{arXiv preprint arXiv:1810.04805}, 15.

\bibitem[{Devlin et~al.(2019)Devlin, Chang, Lee, and Toutanova}]{devlin-etal-2019-bert}
Devlin, J.; Chang, M.-W.; Lee, K.; and Toutanova, K. 2019.
\newblock {BERT}: Pre-training of Deep Bidirectional Transformers for Language Understanding.
\newblock In \emph{Proceedings of the 2019 Conference of the North {A}merican Chapter of the Association for Computational Linguistics: Human Language Technologies, Volume 1 (Long and Short Papers)}, 4171--4186.

\bibitem[{Ding et~al.(2020)Ding, Zhou, Yang, and Tang}]{ding2020cogltx}
Ding, M.; Zhou, C.; Yang, H.; and Tang, J. 2020.
\newblock Cogltx: Applying bert to long texts.
\newblock \emph{Advances in Neural Information Processing Systems}, 33: 12792--12804.

\bibitem[{Galke and Scherp(2022)}]{galke-scherp-2022-bag}
Galke, L.; and Scherp, A. 2022.
\newblock {Bag-of-Words} vs. Graph vs. Sequence in Text Classification: Questioning the Necessity of Text-Graphs and the Surprising Strength of a Wide {MLP}.
\newblock In \emph{Proceedings of the 60th Annual Meeting of the Association for Computational Linguistics (Volume 1: Long Papers)}, 4038--4051.

\bibitem[{Greene and Cunningham(2006)}]{greene2006practical}
Greene, D.; and Cunningham, P. 2006.
\newblock Practical solutions to the problem of diagonal dominance in kernel document clustering.
\newblock In \emph{Proceedings of the 23rd international conference on Machine learning}, 377--384.

\bibitem[{Gu et~al.(2023)Gu, Wang, Zhang, Wu, and Gu}]{gu2023enhancing}
Gu, Y.; Wang, Y.; Zhang, H.-R.; Wu, J.; and Gu, X. 2023.
\newblock Enhancing text classification by graph neural networks with multi-granular topic-aware graph.
\newblock \emph{IEEE Access}, 11: 20169--20183.

\bibitem[{Hassan and Banea(2006)}]{hassan-banea-2006-random}
Hassan, S.; and Banea, C. 2006.
\newblock Random-Walk Term Weighting for Improved Text Classification.
\newblock In \emph{Proceedings of {T}ext{G}raphs: the First Workshop on Graph Based Methods for Natural Language Processing}, 53--60.

\bibitem[{He et~al.(2019)He, Wang, Liu, Feng, and Wu}]{he2019long}
He, J.; Wang, L.; Liu, L.; Feng, J.; and Wu, H. 2019.
\newblock Long document classification from local word glimpses via recurrent attention learning.
\newblock \emph{IEEE Access}, 7: 40707--40718.

\bibitem[{Hu et~al.(2022)Hu, Hosseini, Skorupa~Parolin, Osorio, Khan, Brandt, and D{'}Orazio}]{hu-etal-2022-conflibert}
Hu, Y.; Hosseini, M.; Skorupa~Parolin, E.; Osorio, J.; Khan, L.; Brandt, P.; and D{'}Orazio, V. 2022.
\newblock {C}onfli{BERT}: A Pre-trained Language Model for Political Conflict and Violence.
\newblock In Carpuat, M.; de~Marneffe, M.-C.; and Meza~Ruiz, I.~V., eds., \emph{Proceedings of the 2022 Conference of the North American Chapter of the Association for Computational Linguistics: Human Language Technologies}, 5469--5482. Seattle, United States: Association for Computational Linguistics.

\bibitem[{Huang et~al.(2019)Huang, Ma, Li, Zhang, and Wang}]{huang-etal-2019-text}
Huang, L.; Ma, D.; Li, S.; Zhang, X.; and Wang, H. 2019.
\newblock Text Level Graph Neural Network for Text Classification.
\newblock In \emph{Proceedings of the 2019 Conference on Empirical Methods in Natural Language Processing and the 9th International Joint Conference on Natural Language Processing}, 3444--3450.

\bibitem[{Huang, Chen, and Chen(2022)}]{huang-etal-2022-contexting}
Huang, Y.-H.; Chen, Y.-H.; and Chen, Y.-S. 2022.
\newblock {C}on{T}ext{ING}: Granting Document-Wise Contextual Embeddings to Graph Neural Networks for Inductive Text Classification.
\newblock In \emph{Proceedings of the 29th International Conference on Computational Linguistics}, 1163--1168.

\bibitem[{Kiesel et~al.(2019)Kiesel, Mestre, Shukla, Vincent, Adineh, Corney, Stein, and Potthast}]{kiesel-etal-2019-semeval}
Kiesel, J.; Mestre, M.; Shukla, R.; Vincent, E.; Adineh, P.; Corney, D.; Stein, B.; and Potthast, M. 2019.
\newblock {S}em{E}val-2019 Task 4: Hyperpartisan News Detection.
\newblock In May, J.; Shutova, E.; Herbelot, A.; Zhu, X.; Apidianaki, M.; and Mohammad, S.~M., eds., \emph{Proceedings of the 13th International Workshop on Semantic Evaluation}, 829--839. Minneapolis, Minnesota, USA: Association for Computational Linguistics.

\bibitem[{Kingma and Ba(2014)}]{kingma2014adam}
Kingma, D.~P.; and Ba, J. 2014.
\newblock Adam: A method for stochastic optimization.
\newblock \emph{arXiv preprint arXiv:1412.6980}.

\bibitem[{Li et~al.(2023{\natexlab{a}})Li, Aitken, Bhambhoria, and Zhu}]{li-etal-2023-prefix}
Li, J.; Aitken, W.; Bhambhoria, R.; and Zhu, X. 2023{\natexlab{a}}.
\newblock Prefix Propagation: Parameter-Efficient Tuning for Long Sequences.
\newblock In Rogers, A.; Boyd-Graber, J.; and Okazaki, N., eds., \emph{Proceedings of the 61st Annual Meeting of the Association for Computational Linguistics (Volume 2: Short Papers)}, 1408--1419. Toronto, Canada: Association for Computational Linguistics.

\bibitem[{Li et~al.(2025{\natexlab{a}})Li, Sleem, Gentile, Nichil, and State}]{li2025smalllanguagemodelsreal}
Li, L.; Sleem, L.; Gentile, N.; Nichil, G.; and State, R. 2025{\natexlab{a}}.
\newblock Small Language Models in the Real World: Insights from Industrial Text Classification.
\newblock arXiv:2505.16078.

\bibitem[{Li et~al.(2025{\natexlab{b}})Li, Fu, Chen, and Hu}]{li2025cographnet}
Li, P.; Fu, X.; Chen, J.; and Hu, J. 2025{\natexlab{b}}.
\newblock {CoGraphNet} for enhanced text classification using word-sentence heterogeneous graph representations and improved interpretability.
\newblock \emph{Scientific Reports}, 15(1): 356.

\bibitem[{Li and Aletras(2022)}]{li-aletras-2022-improving}
Li, W.; and Aletras, N. 2022.
\newblock Improving Graph-Based Text Representations with Character and Word Level N-grams.
\newblock In \emph{Proceedings of the 2nd Conference of the Asia-Pacific Chapter of the Association for Computational Linguistics and the 12th International Joint Conference on Natural Language Processing (Volume 2: Short Papers)}, 228--233.

\bibitem[{Li et~al.(2023{\natexlab{b}})Li, Li, Luo, Xie, Lee, Zhao, Wang, and Li}]{li-etal-2023-recurrent}
Li, X.; Li, Z.; Luo, X.; Xie, H.; Lee, X.; Zhao, Y.; Wang, F.~L.; and Li, Q. 2023{\natexlab{b}}.
\newblock Recurrent Attention Networks for Long-text Modeling.
\newblock In Rogers, A.; Boyd-Graber, J.; and Okazaki, N., eds., \emph{Findings of the Association for Computational Linguistics: ACL 2023}, 3006--3019. Toronto, Canada: Association for Computational Linguistics.

\bibitem[{Li et~al.(2024)Li, Wang, Wang, and Wang}]{li2024graph}
Li, X.; Wang, B.; Wang, Y.; and Wang, M. 2024.
\newblock Graph-based text classification by contrastive learning with text-level graph augmentation.
\newblock \emph{ACM Transactions on Knowledge Discovery from Data}, 18(4): 1--21.

\bibitem[{Liu et~al.(2019)Liu, Ott, Goyal, Du, Joshi, Chen, Levy, Lewis, Zettlemoyer, and Stoyanov}]{liu2019roberta}
Liu, Y.; Ott, M.; Goyal, N.; Du, J.; Joshi, M.; Chen, D.; Levy, O.; Lewis, M.; Zettlemoyer, L.; and Stoyanov, V. 2019.
\newblock Roberta: A robustly optimized bert pretraining approach.
\newblock \emph{arXiv preprint arXiv:1907.11692}.

\bibitem[{Liu et~al.(2025)Liu, Huang, Xia, and Zhang}]{liu2025all}
Liu, Z.; Huang, Y.; Xia, X.; and Zhang, Y. 2025.
\newblock All is attention for multi-label text classification.
\newblock \emph{Knowledge and Information Systems}, 67(2): 1249--1270.

\bibitem[{Mihalcea and Tarau(2004)}]{mihalcea-tarau-2004-textrank}
Mihalcea, R.; and Tarau, P. 2004.
\newblock {T}ext{R}ank: Bringing Order into Text.
\newblock In \emph{Proceedings of the 2004 Conference on Empirical Methods in Natural Language Processing}, 404--411.

\bibitem[{Mohammadi and Ghosh(2025)}]{mohammadi2025prototypebasedmodelsetclassification}
Mohammadi, M.; and Ghosh, S. 2025.
\newblock A prototype-based model for set classification.
\newblock arXiv:2408.13720.

\bibitem[{Onan(2023)}]{onan2023hierarchical}
Onan, A. 2023.
\newblock Hierarchical graph-based text classification framework with contextual node embedding and BERT-based dynamic fusion.
\newblock \emph{Journal of King Saud University-Computer and Information Sciences}, 35(7).

\bibitem[{Park, Vyas, and Shah(2022)}]{park-etal-2022-efficient}
Park, H.; Vyas, Y.; and Shah, K. 2022.
\newblock Efficient Classification of Long Documents Using Transformers.
\newblock In Muresan, S.; Nakov, P.; and Villavicencio, A., eds., \emph{Proceedings of the 60th Annual Meeting of the Association for Computational Linguistics (Volume 2: Short Papers)}, 702--709. Dublin, Ireland: Association for Computational Linguistics.

\bibitem[{Pennington, Socher, and Manning(2014)}]{pennington2014glove}
Pennington, J.; Socher, R.; and Manning, C.~D. 2014.
\newblock Glove: Global vectors for word representation.
\newblock In \emph{Proceedings of the 2014 conference on empirical methods in natural language processing (EMNLP)}, 1532--1543.

\bibitem[{Raffel et~al.(2020)Raffel, Shazeer, Roberts, Lee, Narang, Matena, Zhou, Li, and Liu}]{raffel2020exploring}
Raffel, C.; Shazeer, N.; Roberts, A.; Lee, K.; Narang, S.; Matena, M.; Zhou, Y.; Li, W.; and Liu, P.~J. 2020.
\newblock Exploring the limits of transfer learning with a unified text-to-text transformer.
\newblock \emph{Journal of machine learning research}, 21(140): 1--67.

\bibitem[{Reusens et~al.(2024)Reusens, Stevens, Tonglet, {De Smedt}, Verbeke, {vanden Broucke}, and Baesens}]{REUSENS2024124302}
Reusens, M.; Stevens, A.; Tonglet, J.; {De Smedt}, J.; Verbeke, W.; {vanden Broucke}, S.; and Baesens, B. 2024.
\newblock Evaluating text classification: A benchmark study.
\newblock \emph{Expert Systems with Applications}, 254: 124302.

\bibitem[{Rousseau, Kiagias, and Vazirgiannis(2015)}]{rousseau-etal-2015-text}
Rousseau, F.; Kiagias, E.; and Vazirgiannis, M. 2015.
\newblock Text Categorization as a Graph Classification Problem.
\newblock In \emph{Proceedings of the 53rd Annual Meeting of the Association for Computational Linguistics and the 7th International Joint Conference on Natural Language Processing (Volume 1: Long Papers)}, 1702--1712.

\bibitem[{Sahu et~al.(2019)Sahu, Christopoulou, Miwa, and Ananiadou}]{sahu-etal-2019-inter}
Sahu, S.~K.; Christopoulou, F.; Miwa, M.; and Ananiadou, S. 2019.
\newblock Inter-sentence Relation Extraction with Document-level Graph Convolutional Neural Network.
\newblock In Korhonen, A.; Traum, D.; and M{\`a}rquez, L., eds., \emph{Proceedings of the 57th Annual Meeting of the Association for Computational Linguistics}, 4309--4316. Florence, Italy: Association for Computational Linguistics.

\bibitem[{Singh et~al.(2022)Singh, Devi, Devi, and Mahanta}]{SINGH2022100061}
Singh, K.~N.; Devi, S.~D.; Devi, H.~M.; and Mahanta, A.~K. 2022.
\newblock A novel approach for dimension reduction using word embedding: An enhanced text classification approach.
\newblock \emph{International Journal of Information Management Data Insights}, 2(1): 100061.

\bibitem[{Touvron et~al.(2023)Touvron, Lavril, Izacard, Martinet, Lachaux, Lacroix, Rozi{\`e}re, Goyal, Hambro, Azhar et~al.}]{touvron2023llama}
Touvron, H.; Lavril, T.; Izacard, G.; Martinet, X.; Lachaux, M.-A.; Lacroix, T.; Rozi{\`e}re, B.; Goyal, N.; Hambro, E.; Azhar, F.; et~al. 2023.
\newblock Llama: Open and efficient foundation language models.
\newblock \emph{arXiv preprint arXiv:2302.13971}.

\bibitem[{Veli{\v{c}}kovi{\'c} et~al.(2017)Veli{\v{c}}kovi{\'c}, Cucurull, Casanova, Romero, Lio, and Bengio}]{velivckovic2017graph}
Veli{\v{c}}kovi{\'c}, P.; Cucurull, G.; Casanova, A.; Romero, A.; Lio, P.; and Bengio, Y. 2017.
\newblock Graph attention networks.
\newblock \emph{arXiv preprint arXiv:1710.10903}.

\bibitem[{Wang et~al.(2023)Wang, Wang, Zhan, Ma, and Jiang}]{wang2023text}
Wang, Y.; Wang, C.; Zhan, J.; Ma, W.; and Jiang, Y. 2023.
\newblock Text FCG: Fusing contextual information via graph learning for text classification.
\newblock \emph{Expert Systems with Applications}, 219: 119658.

\bibitem[{Warner et~al.(2024)Warner, Chaffin, Clavi{\'e}, Weller, Hallstr{\"o}m, Taghadouini, Gallagher, Biswas, Ladhak, Aarsen et~al.}]{warner2024smarter}
Warner, B.; Chaffin, A.; Clavi{\'e}, B.; Weller, O.; Hallstr{\"o}m, O.; Taghadouini, S.; Gallagher, A.; Biswas, R.; Ladhak, F.; Aarsen, T.; et~al. 2024.
\newblock Smarter, better, faster, longer: A modern bidirectional encoder for fast, memory efficient, and long context finetuning and inference.
\newblock \emph{arXiv preprint arXiv:2412.13663}.

\bibitem[{Wortsman et~al.(2023)Wortsman, Lee, Gilmer, and Kornblith}]{wortsman2023replacing}
Wortsman, M.; Lee, J.; Gilmer, J.; and Kornblith, S. 2023.
\newblock Replacing softmax with relu in vision transformers.
\newblock \emph{arXiv preprint arXiv:2309.08586}.

\bibitem[{Xu et~al.(2021)Xu, Chen, Ma, Huang, and Xiang}]{xu-etal-2021-contrastive-document}
Xu, P.; Chen, X.; Ma, X.; Huang, Z.; and Xiang, B. 2021.
\newblock Contrastive Document Representation Learning with Graph Attention Networks.
\newblock In \emph{Findings of the Association for Computational Linguistics: EMNLP 2021}, 3874--3884.

\bibitem[{Yang et~al.(2019)Yang, Dai, Yang, Carbonell, Salakhutdinov, and Le}]{yang2019xlnet}
Yang, Z.; Dai, Z.; Yang, Y.; Carbonell, J.; Salakhutdinov, R.~R.; and Le, Q.~V. 2019.
\newblock Xlnet: Generalized autoregressive pretraining for language understanding.
\newblock \emph{Advances in Neural Information Processing Systems}, 32.

\bibitem[{Yao, Mao, and Luo(2019)}]{Yao_Mao_Luo_2019}
Yao, L.; Mao, C.; and Luo, Y. 2019.
\newblock Graph Convolutional Networks for Text Classification.
\newblock \emph{Proceedings of the AAAI Conference on Artificial Intelligence}, 33: 7370--7377.

\bibitem[{Yun, Kim, and Kim(2023)}]{yun-etal-2023-focus}
Yun, J.; Kim, M.; and Kim, Y. 2023.
\newblock Focus on the Core: Efficient Attention via Pruned Token Compression for Document Classification.
\newblock In Bouamor, H.; Pino, J.; and Bali, K., eds., \emph{Findings of the Association for Computational Linguistics: EMNLP 2023}, 13617--13628. Singapore: Association for Computational Linguistics.

\bibitem[{Zhang et~al.(2020)Zhang, Yu, Cui, Wu, Wen, and Wang}]{zhang-etal-2020-every}
Zhang, Y.; Yu, X.; Cui, Z.; Wu, S.; Wen, Z.; and Wang, L. 2020.
\newblock Every Document Owns Its Structure: Inductive Text Classification via Graph Neural Networks.
\newblock In Jurafsky, D.; Chai, J.; Schluter, N.; and Tetreault, J., eds., \emph{Proceedings of the 58th Annual Meeting of the Association for Computational Linguistics}, 334--339. Online: Association for Computational Linguistics.

\bibitem[{Zhao et~al.(2024)Zhao, Xu, Xiao, Jia, and Hou}]{zhao2024mobilediffusion}
Zhao, Y.; Xu, Y.; Xiao, Z.; Jia, H.; and Hou, T. 2024.
\newblock {MobileDiffusion}: Instant text-to-image generation on mobile devices.
\newblock In \emph{European Conference on Computer Vision}, 225--242. Springer.

\bibitem[{Zhu et~al.(2015)Zhu, Kiros, Zemel, Salakhutdinov, Urtasun, Torralba, and Fidler}]{zhu2015aligning}
Zhu, Y.; Kiros, R.; Zemel, R.; Salakhutdinov, R.; Urtasun, R.; Torralba, A.; and Fidler, S. 2015.
\newblock Aligning books and movies: Towards story-like visual explanations by watching movies and reading books.
\newblock In \emph{Proceedings of the IEEE international conference on computer vision}, 19--27.

\end{thebibliography}

\appendix
\section{Graph-Based vs. Non-Graph Approaches}
\label{appendix:nongraph}

\subsection{Classification Methods}
\begin{table*}[!t]
    \centering    
    \begin{tabularx}{\textwidth}{p{6.7cm}RRRRRR}
        \toprule
        & \multicolumn{2}{c}{\textbf{BBC News}} & \multicolumn{2}{c}{\textbf{HND}} & \multicolumn{2}{c}{\textbf{arXiv}} \\
        \cmidrule(lr){2-3} \cmidrule(lr){4-5} \cmidrule(lr){6-7}
        \textbf{Graph Scheme} & \textbf{Accuracy} & \textbf{$\mathbf{F_1}$-ma} & \textbf{Accuracy} & \textbf{$\mathbf{F_1}$-ma} & \textbf{Accuracy} & \textbf{$\mathbf{F_1}$-ma} \\
        \hline
        \textit{Non-graph-based strategies} \\
        \hline
        Longformer \cite{park-etal-2022-efficient} & -- & -- & 95.7 & -- & -- & -- \\
        BERT \cite{park-etal-2022-efficient} & -- & -- & 92.0 & -- & -- & -- \\
        CogLTX \cite{park-etal-2022-efficient} & -- & -- & 94.8 & -- & -- & -- \\
        rRF \cite{SINGH2022100061} & 96.2 & 96.1 & -- &-- &-- &-- \\
        ConfliBERT-SCR \cite{hu-etal-2022-conflibert} & -- & 98.1 & -- &-- &-- &-- \\
        Prefix-Propagation \cite{li-etal-2023-prefix} & -- & -- & -- & 81.8 & -- & 83.3 \\
        LSG \cite{10.1007/978-3-031-33374-3_35} & -- & -- & -- & -- & -- & 87.9 \\
        RAN+Random \cite{li-etal-2023-recurrent} & -- & -- & 93.9 & -- & 80.1 & -- \\
        RAN+GloVe \cite{li-etal-2023-recurrent} & -- & -- & 95.4 & -- &  83.4 & -- \\
        RAN+Pretrain \cite{li-etal-2023-recurrent} & -- & -- & $\mathbf{96.9}$ & -- & 85.9 & -- \\
        PFC \cite{yun-etal-2023-focus} & 98.1 & 97.1 & -- &	-- & $\star$76.0 & $\star$61.0 \\
        RoBERTa \cite{REUSENS2024124302} & 98.0 & 97.0 & -- &-- & -- & -- \\
        Llama-3.2-1B-Instruct \cite{li2025smalllanguagemodelsreal} & -- & -- & -- & -- & 89.2 & 89.0 \\
        Llama-3.2-3B-Instruct \cite{li2025smalllanguagemodelsreal} & -- & -- & -- & -- & 90.4 & 90.3 \\
        ModernBERT-base \cite{li2025smalllanguagemodelsreal} & -- & -- & -- & -- & 81.0 & 81.1 \\
        AChorDS-LVQ \cite{mohammadi2025prototypebasedmodelsetclassification} & -- & -- & 91.8 & -- & -- & --\\
        \hline 
        \textit{Heuristic-based graphs} \\
        \hline
        complete graph &  $\mathbf{99.9}$ & $\mathbf{99.9}$ & 94.6 & 94.5 & -- & -- \\
        sentence order & 99.7 & 99.7 & 92.6 & 92.6 & 87.3 & 86.7 \\
        window co-occurrence & 99.8 & 99.8 & 92.1 & 92.1 & 87.9 & 87.4 \\
        mean semantic similarity & 99.4 & 99.3 & 91.2 & 91.1 & -- & --\\
        max semantic similarity & 99.7 & 99.7 & 92.8 & 92.8 & 87.8 & 87.4 \\
        \hline
        \textit{Our learned graphs} \\
        \hline 
        learned mean-bound & $\mathbf{99.9}$ & $\mathbf{99.9}$ & 95.0 & $\mathbf{94.9}$ & -- & -- \\ 
        learned max-bound & 99.6 & 99.6 &  92.6 & 92.6 & $\mathbf{91.9}$ & $\mathbf{91.7}$ \\               
        \bottomrule        
    \end{tabularx}
        \caption{Classification results of proposed learned graph structures compared to heuristic-based graph construction methods and recent non-graph-based approaches. Reported metrics include accuracy and macro-averaged $F_1$ score for each dataset. Notably, the results marked with $\star$ are not comparable to the models here reported, as the corresponding authors used a subsample of the arXiv dataset and performed the classification based on the abstract of the articles as the input.}
    \label{tab:nongraph_results}
\end{table*}

While the focus of this work is on graph-based strategies for document representation and their impact on document classification tasks, we also provide a comparative overview of recent non-graph-based approaches utilizing traditional vector-based representations for document classification. Table \ref{tab:nongraph_results} summarizes the performance of recently proposed models on the datasets considered in this paper.  

\cite{park-etal-2022-efficient} fine-tuned several Transformer-based models including \textbf{BERT} \cite{devlin2018bert}, \textbf{Longformer} \cite{beltagy2020longformer}, and \textbf{CogLTX} \cite{ding2020cogltx}. BERT was finetuned on truncated inputs to the first 512 tokens, using a fully-connected layer on the [CLS] token for classification. Longformer, which supports longer input sequences (up to 4,096 tokens) via sparse self-attention, also utilized a fully connected layer on top of the [CLS] token with global attention for the classification. The Cognize Long TeXts (CogLTX) model was also included in the study with the hypothesis that a small set of key sentences is sufficient for accurate document classification. 

Another method, \textbf{rRF} (removal of Redundant Feature) \cite{SINGH2022100061} applies dimensionality reduction by eliminating redundant information based on word-level similarity scores computed using GloVe embeddings \cite{pennington2014glove}, followed by a Naive Bayes classifier. 

\textbf{ConfliBERT} \cite{hu-etal-2022-conflibert} is a domain-specific pre-trained language model
for conflict and political violence detection. Although the authors explore both pretraining from scratch and continual pretraining strategies, Table \ref{tab:nongraph_results} only reports the best-performing variant -- pretrained from scratch using cased data (SCR). 

Although parameter-efficient tuning methods aim to reduce memory overhead while attaining comparable performance to fine-tuning of pretrained language models, they often fail to model long documents. To address this, \cite{li-etal-2023-prefix} propose \textbf{Prefix-Propagation}, a technique that allows prefix hidden states to dynamically evolve across layers by incorporating them into the attention mechanism.  

To further mitigate the quadratic complexity of Transformer self-attention for long sequences, Local Sparse Global (\textbf{LSG}) attention is proposed in \cite{10.1007/978-3-031-33374-3_35}. LSG follows a block-based processing of the input and applies local attention to capture local context for nearby dependencies, sparse attention for extended context, and global attention to improve information flow inside the model.

In a similar direction, \cite{li-etal-2023-recurrent} propose the Recurrent Attention Network (\textbf{RAN}),  which introduces a recurrent formulation of self-attention to handle long sequences, enabling long-term memory and extracting global semantics in both token-level and document-level representations. RAN processes sequences in non-overlapping windows, applying positional multi-head self-attention to a window area, and propagates a global perception cell vector across windows to capture long-term dependencies. 
Table \ref{tab:nongraph_results} presents results for three RAN variants: i) RAN+Random, with randomly initialized weights; ii) RAN+GloVe, using GloVe embedding \cite{pennington2014glove} as word representation; and iii) RAN+Pretrain, pretrained with a masked language modeling objective on the BookCorpus \cite{zhu2015aligning} and C4 (RealNews-like subset) \cite{raffel2020exploring}.

To further reduce the computation of self-attention, \cite{yun-etal-2023-focus} propose a \textbf{PFC} strategy, which integrates a token pruning step to eliminate less important tokens from attention computations, and a token combining step to condense input sequences into smaller sizes.

Despite such innovations, full model fine-tuning remains widely adopted in document classification. For instance, a fine-tuned \textbf{RoBERTa} \cite{liu2019roberta} was used in \cite{REUSENS2024124302}, combining Bayesian search with author recommendations for hyperparameter setting. Similarly, \cite{li2025smalllanguagemodelsreal} evaluate small language models in real-world classification tasks, focusing on best practices and tuning strategies to address text classification effectively. The study included \textbf{Llama3.2 (1B-3B)} \cite{touvron2023llama} and \textbf{ModernBERT-base} \cite{warner2024smarter}.

Finally, Adaptive Chordal Distance and Subspace-based LVQ (\textbf{AChorDS-LVQ}) \cite{mohammadi2025prototypebasedmodelsetclassification} is introduced as a prototype-based approach for learning on the manifold of linear subspaces derived from input vectors. The method learns a set of subspace prototypes to represent class characteristics and relevance factors, automating the selection of subspace dimensionalities and the influence of each input vector on classification outcomes.

\subsection{Observed Results}

In both the BBC News and arXiv datasets, our learned graph structures consistently outperform all baseline models, including both heuristic-based graphs and recent non-graph approaches.
On BBC News, our learned mean-bound graphs achieve near-perfect performance with 99.9\% accuracy and $F_1$ score, significantly surpassing the best non-graph alternative, PFC, which reaches 98.1\% accuracy and 97.1\% $F_1$ score. Similarly, on arXiv, our learned max-bound graphs have a considerable advantage over other graphs as well as over the strongest non-graph model, fine-tuned Llama-3.2. While Llama-3.2 reports 90.4\% accuracy for the 3B version and 89.2\% accuracy for the 1B variant, our learned graphs yield 91.9\% accuracy and 91.7\% $F_1$ score without requiring manual constructions or task-specific expert knowledge.  
In contrast, on the HND dataset, heuristic-based graph methods underperform compared to non-graph baselines.
However, our learned graphs remain competitive with the top-performing models, such as RAN and fine-tuned Longformer and CogLTX, demonstrating the capacity of our learned graphs to capture the document structure. 

The observed results underscore the effectiveness of automatically identifying task-relevant segments within input sequences, supporting the integration of local contextual information at lower textual granularities while preserving global semantics at higher levels. Moreover, the performance of RAN demonstrates the benefit of attention mechanisms that operate over windows with explicit propagation of information from fine-grained units (e.g., tokens) to higher-level representations. Such a strategy offers a clear advantage over conventional sequential models in constructing comprehensive document representations.
The results from Table \ref{tab:nongraph_results} further motivate future work to explore alternative filtering strategies, other attention mechanisms, and hierarchical approaches to constructing graphs over multiple text granularities (e.g., sentences, sections) via heterogeneous graph structures.

\end{document}